\documentclass[lettersize,journal]{IEEEtran}
\usepackage{amsmath,amsfonts}
\usepackage{algorithmic}
\usepackage{algorithm}
\usepackage{array}
\usepackage[caption=false,font=normalsize,labelfont=sf,textfont=sf]{subfig}
\usepackage{textcomp}
\usepackage{stfloats}
\usepackage{url}
\usepackage{verbatim}
\usepackage{graphicx}
\usepackage{cite}
\usepackage{utfsym,colortbl}
\begin{document}

\title{STL: A Signed and Truncated Logarithm Activation Function for Neural Networks}

\author{Yuanhao Gong\\College of Electronics and Information Engineering, Shenzhen University, China.~~gong.ai@qq.com}

\markboth{Journal of \LaTeX\ Class Files,~Vol.~14, No.~8, August~2021}%
{Shell \MakeLowercase{\textit{et al.}}: A Sample Article Using IEEEtran.cls for IEEE Journals}


\maketitle

\begin{abstract}
Activation functions play an essential role in neural networks. They provide the non-linearity for the networks. Therefore, their properties are important for neural networks' accuracy and running performance. In this paper, we present a novel signed and truncated logarithm function as activation function. The proposed activation function has significantly better mathematical properties, such as being odd function, monotone, differentiable, having unbounded value range, and a continuous nonzero gradient. These properties make it an excellent choice as an activation function. We compare it with other well-known activation functions in several well-known neural networks. The results confirm that it is the state-of-the-art. The suggested activation function can be applied in a large range of neural networks where activation functions are necessary.
\end{abstract}


\section{Introduction}
\IEEEPARstart{A}{ctivation} functions are a vital component of neural networks that play a crucial role in their success. They help to introduce non-linearity into the output of a neural network and make it capable of modeling complex relationships between inputs and outputs. Without activation functions, a neural network would be limited to linear operations, and complex patterns would be challenging to identify. Their introduced non-linearity empowers it to learn intricate patterns that would be otherwise impossible to learn.

There are several activation functions available, such as Sigmoid and ReLU, each with its unique strengths and weaknesses. By selecting the appropriate activation function for the neural network being developed, designers can optimize the network's performance and ensure its success. For instance, the $sigmoid$ function is ideal for binary classification tasks since it returns a value between 0 and 1, which makes it easy to interpret. In contrast, the $ReLU$ function is highly efficient at handling sparse input data, making it particularly useful in deep neural networks~\cite{Nair2010}. Similarly, the $tanh$ function produces output values between -1 and 1, making it an excellent choice for data that has negative values. More details are in Section~\ref{sec:work}.

It is crucial to note that selecting the right activation function is not a one-size-fits-all approach. The choice of the activation function depends on several factors, such as the type of data being used, the specific requirements of the neural network, and the nature of the problem being solved. By carefully selecting the appropriate activation function, designers can enhance the neural network's performance, making it more accurate and efficient in identifying complex patterns.
\begin{figure}
	\centering
		\includegraphics[width=0.45\linewidth]{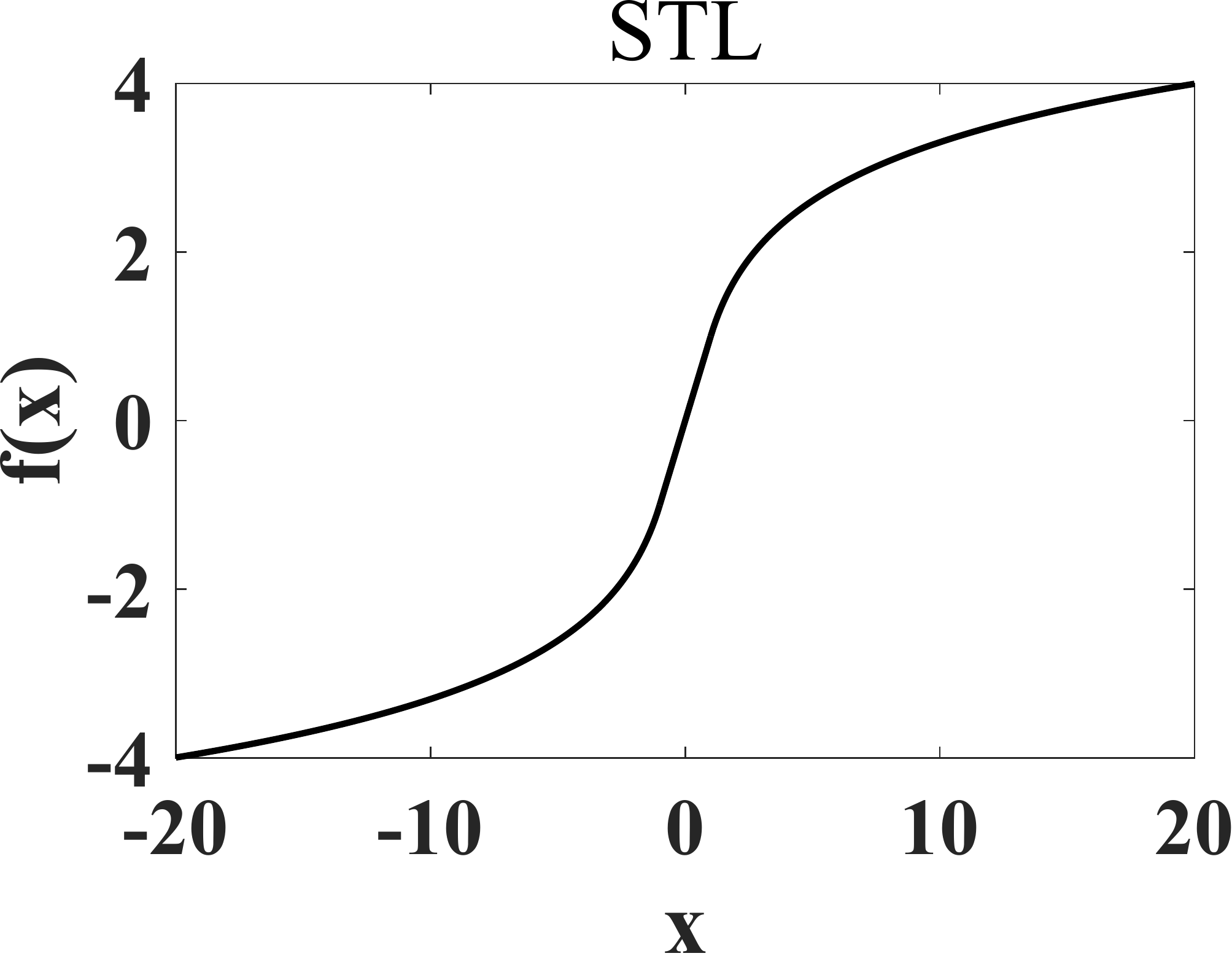}
		~~~
		\includegraphics[width=0.45\linewidth]{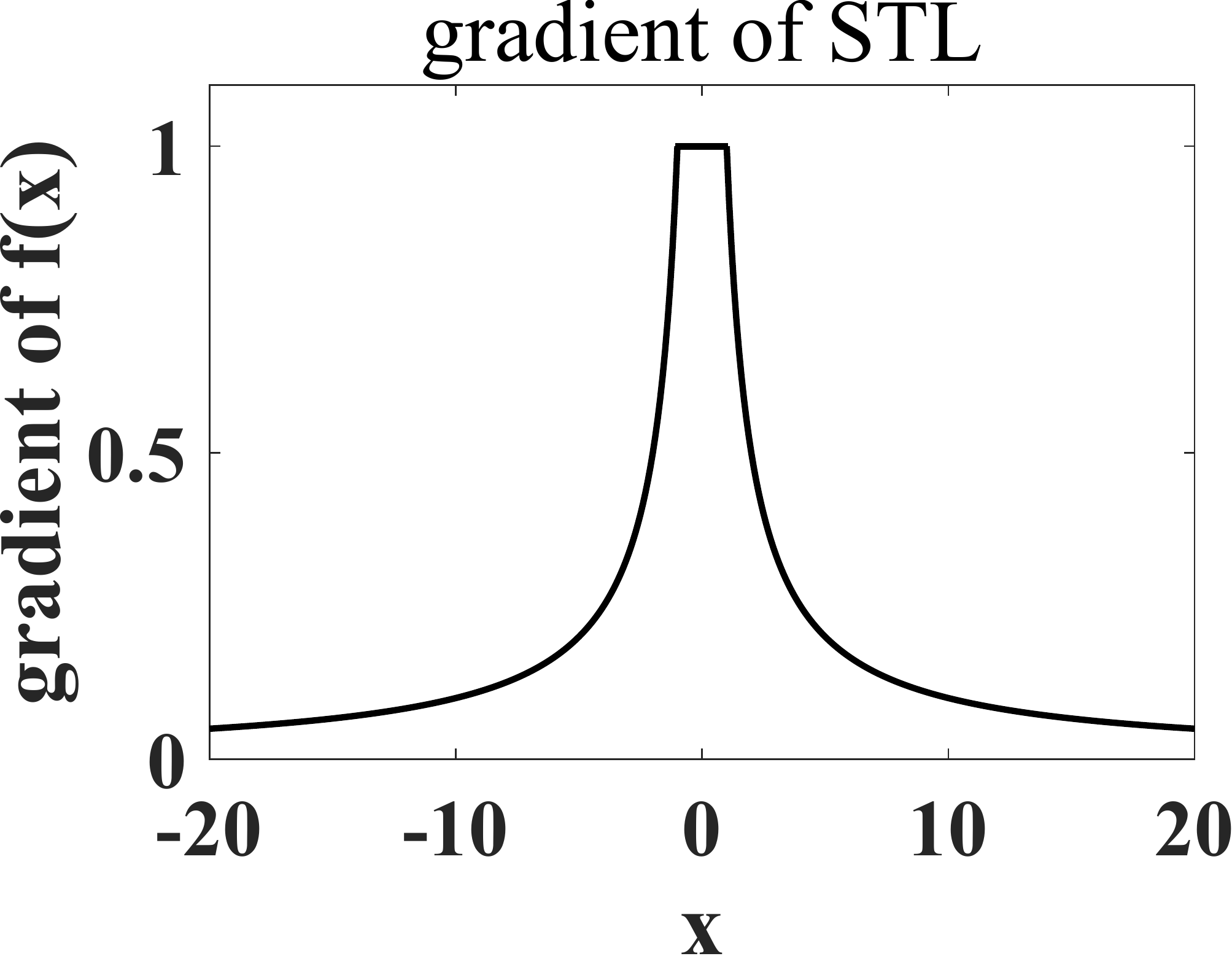}
	\caption{The proposed activation function (left) and its gradient (right) with parameter $\alpha=1$. This activation function has several desired properties such as being odd, monotone, differentiable and having positive gradient. Its gradient function has a $\pi$ shape and is always positive.}
	\label{fig1}
\end{figure}
\subsection{Related Work}
\label{sec:work}
There are some well-known activation functions available. The famous $Sigmoid$ function is defined as
\begin{equation}
	f_1(x)=\frac{1}{1+\exp(-x)}\,,
\end{equation} whose values are bounded in $(0,1)$.
The popular $ReLU$ is defined as
\begin{equation}
	f_2(x)=\max(x,0)\,,
\end{equation} whose values are bounded in $(0,+\infty)$. It has some variants such as $PReLU$, which is defined as
\begin{equation}
	f_3(x)=\left\{
	\begin{array}{ll}
		x, &\mathrm{when }\, x>0 \\
		\alpha x, &\mathrm{else}
	\end{array}\right.,
\end{equation} whose values are not bounded. Another variant $ELU$ is
\begin{equation}
	f_4(x)=\left\{
	\begin{array}{ll}
		x, &\mathrm{when }\, x>0 \\
		\alpha (e^x-1), &\mathrm{else}
	\end{array}\right..
\end{equation} Another activation $swish$ is defined as
\begin{equation}
	f_5(x)=x*Sigmoid(x)\,,
\end{equation} which has a negative lower bound. The $\tanh$ function is 
\begin{equation}
	f_6(x)=\frac{e^{x}-e^{-x}}{e^{x}+e^{-x}}\,,
\end{equation} whose values are in $(-1,1)$. The $softsign$ function is
\begin{equation}
	f_7(x)=\frac{x}{|x|+1}\,,
\end{equation}whose value are also in $(-1,1)$. The well-known $softmax$ is
\begin{equation}
	f_8(x_i)=\frac{e^{x_i}}{\sum_{i}e^{x_i}}\,,
\end{equation} which is frequently used in classification tasks. Very recently, $f_8$ has been shown to lead to numerical issues in~\cite{Bondarenko2023}. An activation function in~\cite{Liu2019b} named the natural logarithm rectified linear unit (NLReLU) is defined as
\begin{equation}
	f_9(x)=\log(\alpha*\max(x,0)+1)\,,
\end{equation} which is non negative. Recently, an activation function named Serf is defined as~\cite{Nag2023}
\begin{equation}
	f_{10}(x)=x*\mathrm{erf}(\log(e^x+1))\,,
\end{equation} where $erf$ is the error function.

There are more activation functions and we can not list all of them. And researchers might design their own activation functions for specific tasks. In the following subsection, we analyze the mathematical properties of these functions and discuss their limitations.
\subsection{Analysis}
As mentioned, these activation functions provide the non linearity for the neural networks. Therefore, their behavior is fundamentally important for the networks.

There are two types of activation functions. One is centered at the origin, $f(0)=0$, and the other is $f(x)\ge 0$. They correspond to different tasks such as regression and classification. 

In general, a good activation function has to satisfy several properties. And we list some of them as the following.
\subsubsection{odd function}
We believe that the activation function should be an odd function, $f(-x)=-f(x)$ and $f(0)=0$. From mathematical point of view, it is strange to prefer the positive values than the negative values. And such preference might implicitly bias the learning system. (Please distinguish the bias with the bias in the linear transformation). To eliminate such bias, it is better to have an odd function as an activation function. 
\subsubsection{monotone function}
\begin{figure}
	\centering
	\includegraphics[width=0.5\linewidth]{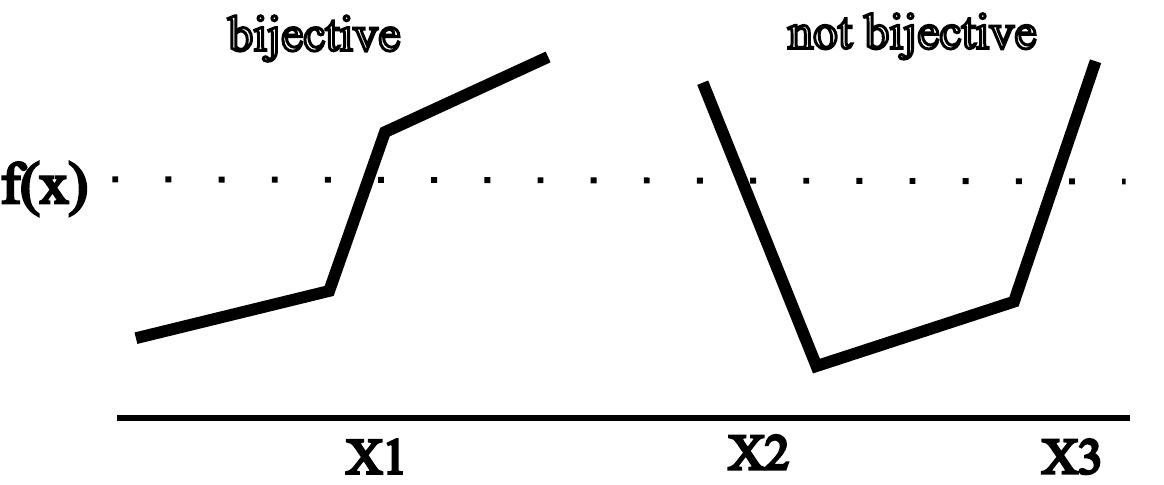}
	\caption{The monotone function is a bejective function and thus the output $f(x)$ has a unique correspondence input $x_1$. In contrast, a non-bijective function might lose such uniqueness ($f(x_2)=f(x_3)$).}
	\label{fig2}
\end{figure}

We also believe that the activation function should be monotone and, in most of cases, non-decreasing. Such property preserves the order in the input. In other words, the larger input is non-linearly mapped into a larger output. This order preserving property is desired. And a monotone function usually is a bijective mapping, which means the output does not lose the information from the input. One example is illustrated in Fig.~\ref{fig2}. Be aware that the distance between two inputs is non-linearly scaled in the output. 
\subsubsection{differentiable}
Another property that an activation function should have is differentiable. With such property, the gradient of the activation function is continuous. Thus the gradient function has no dramatic change in a small neighborhood. The continuity of the gradient function guarantees the numerical stability when performing the back-propagation algorithm.
\subsubsection{unbounded value}
The value of an activation function should fully fill the interval $(-\infty,+\infty)$. In contrast, the function with bounded values such as $softsign$ will have small difference when two inputs have large values. For example, $softsign(1000)\approx 1\approx softsign(2000)$, although $1000$ and $2000$ have a significant numerical difference. In other words, the $softsign$ activation function can not distinguish the two input $1000$ and $2000$, showing its limitations. Similar thing happens for two negative inputs in the $ReLU$ function.
\subsubsection{continuous gradient}
On the other hand, the gradient of the activation function should be continuous and nonzero. The zero gradient (also known as vanishing gradient) is problematic when the back-propagation algorithm is performed. According the monotone property, we expect the gradient is continuous. The continuity guarantees that there is no dramatic change in a small neighborhood. It helps in improving the numerical stability of the neural networks.
\subsubsection{computation efficiency}
The computation efficiency must be token into account because the activation function and its gradient would be evaluated many time during the training and inference process for the neural networks. Therefore, the running time of the networks can be effected by the activation function's computation efficiency.

We consider these six properties as desired properties of activation functions. And we will evaluate previous activation functions in these six aspects.
\subsection{Motivation and Contribution}
Previous activation functions can not satisfy all of the above six aspects. This motivates us to construct a novel function that satisfies these rules.

Our contributions include the following
\begin{itemize}
	\item  we present a novel activation function, which is odd, monotone, differentiable, has unbounded values and bounded gradients. The function and its gradient can be efficiently evaluated.
	\item we analyze the properties of this function and argue why it is preferred as activation function.
	\item we numerically confirm that it performs better than others for many well-known neural networks.
\end{itemize}
\section{Signed and Truncated Logarithm Function}
In this section, we present an activation function that fully satisfies the above six preferred properties. More specifically, we define an activation function as
\begin{equation}
	\label{eq:ours}
	f_{our}(x)=\left\{
	\begin{array}{ll}
		\alpha x, &\mathrm{when }\, |x|\le 1 \\
		\alpha \delta(x)(\log(|x|)+1), &\mathrm{else}
	\end{array}\right.,
\end{equation} where $\delta(x)$ is the sign of $x$ (if $x>0$, $\delta(x)=1$; if $x<0$, $\delta(x)=-1$). The scalar parameter $\alpha>0$ usually is set to $1$. Since the activation function usually is followed by a linear transformation, the value of $\alpha$ is not important. It will be automatically adjusted by the learnable parameters. The $\log$ function is truncated when $|x|\le 1$ because its gradient is dramatically increasing in that interval. Such truncation avoids the numerical issue around the origin. 

\begin{table*}[!t]
	\caption{Comparison the activation functions in the six properties\label{tab:table1}}
	\centering
	\begin{tabular}{c|c|c|c|c|c|c|c}
		\hline
		function & odd & monotone & differentiable & unbounded value range & continuous gradient& \multicolumn{2}{c}{compute time (left) and gradient compute time (right)}\\
		\hline
		$f_1$ & $\usym{2717}$ & $\checkmark$ & $\checkmark$ & $\usym{2717}$& $\checkmark$ & high & high\\
		\hline
		$f_2$ & $\usym{2717}$ & $\checkmark$ & $\usym{2717}$ & lower bound& $\usym{2717}$ & low & low\\
		\hline
		$f_3$ & $\usym{2717}$ & $\checkmark$ & $\usym{2717}$ & $\checkmark$& $\usym{2717}$ & low & low\\
		\hline
		$f_4$ & $\usym{2717}$ & $\checkmark$ & $\checkmark$ & lower bound& $\usym{2717}$ & medium & medium\\
		\hline
		$f_5$ & $\usym{2717}$ & $\usym{2717}$ & $\checkmark$ & lower bound& $\checkmark$ & high & high\\
		\hline
		$f_6$ & $\checkmark$ & $\checkmark$ & $\checkmark$ & $\usym{2717}$& $\checkmark$ & high & high\\
		\hline
		$f_7$ & $\checkmark$ & $\checkmark$ & $\checkmark$ & $\usym{2717}$& $\checkmark$ & low & low\\
		\hline
		$f_8$ & $\usym{2717}$ & $\usym{2717}$ & $\checkmark$ & $\usym{2717}$& $\checkmark$ & high & high\\
		\hline
		$f_9$ & $\usym{2717}$ & $\checkmark$ & $\usym{2717}$ & lower bound& $\usym{2717}$ & low & low\\
		\hline
		$f_{10}$ & $\usym{2717}$ & $\usym{2717}$ & $\checkmark$ & lower bound & $\checkmark$ & high & high\\
		\hline
		$f_{our}$ & $\checkmark$ & $\checkmark$ & $\checkmark$ & $\checkmark$& $\checkmark$ & low & low\\
		\hline
	\end{tabular}
\end{table*}
The gradient of this signed and truncated function is
\begin{equation}
	\label{eq:ourg}
	f'_{our}(x)=\left\{
	\begin{array}{ll}
		\alpha, &\mathrm{when }\, |x|\le 1 \\
		\frac{\alpha }{|x|}, &\mathrm{else}
	\end{array}\right..
\end{equation} It follows that $0<f'_{our}(x)\le \alpha$. Therefore, the gradient never vanishes. The gradient $f'_{our}(x)$ is also continuous, showing its numerical stability. The continuity indicates that there is no dramatic change in a small neighborhood.

We name this function as Signed and Truncated Logarithm (STL) function. And we set the scale parameter $\alpha=1$ by default. This function and its gradient are illustrated in Fig.~\ref{fig1}.

\subsection{Mathematical Properties}
The proposed STL satisfies the preferred six properties in the previous section.
\begin{itemize}
\item It is not difficult to show that STL is odd, $f_{our}(-x)=-f_{our}(x)$. Such property guarantees that there is no bias from the activation function itself. 
\item STL is increasing (monotone) because its gradient is always positive. Therefore, it is a bijective mapping. The monotone bijective mapping is important from information point of view. It means that there is no information collapsed or generated. And the relative order from the input is also preserved in the output. 
\item STL is differentiable. Such smoothness guarantees that there is no dramatic change in a small neighborhood. 
\item STL has unbounded value range. This means that STL can still distinguish two input values even they are large. 
\item Its gradient has bounded value range. Its nonzero gradient property guarantees that the vanishing gradient issue is not caused by the activation function itself, improving the networks' numerical stability. 
\item STL and its gradient can be efficiently computed.
\end{itemize}
\subsection{Computation Properties}
Another advantage of the proposed STL is its numerical efficiency. STL and its gradient can be efficiently computed. For example, when $|x|>1$, STL can be approximated by the $\log_2$ function
\begin{equation}
	f_{our}=\alpha\delta(x)(\frac{\log_2(|x|)}{\log_2(e)}+1)=\beta\log_2(|x|)+ \alpha\delta(x)\,,
\end{equation} where $\beta=\alpha\delta(x)/\log_2(e)$ is a signed constant. The $\log_2$ function has a fast approximation scheme, thanks to the binary representation of $x$.

More specifically, a 32bit float number $x$ can be represented as a sign bit $s$, a 8bit $E$ and the fraction $V$,
\begin{equation}
	x=(-1)^s*2^{E-127}*(1+V)\,.
\end{equation} Therefore, the $\log_2(x)$ (since $x>0$ we have $s=0$) is
\begin{equation}
	\label{eq:acc}
	\begin{split}
	\log_2(x)&=\log_2(1)+E-127+\log_2(1+V)\,\\
			&=E-127+\log_2(1+V)\,.
	\end{split}
\end{equation}
The $E$ is easy to get. We only need to approximate the values of $\log_2(x)$ where $1\le x<2$ via polynomials, for example, $(-0.344845x+2.024658)x-1.674873$, and we can also store these values in a lookup table to further accelerate the computation.

More over, the gradient of STL can also be efficiently computed via Eq.~\ref{eq:ourg}. The gradient is decreasing when $|x|$ is increasing. And their multiplication $|x|*f'_{our}(x)=\alpha$ is a constant when $|x|>1$. 
\subsection{The Scale Parameter $\alpha$}
If we fixed the scale parameter $\alpha=1$, it will not affect the performance of the neural network too much because the activation function usually is followed by a linear transformation which can absorb this scale parameter. Therefore, the value of this parameter is not so important in practice. It usually is set to $1$ for simplicity.

But if we let $0<\alpha<1$, we can theoretically prove that the proposed STL is a contraction mapping ($|f'_{our}(x)|<1$), which is a necessary condition to find the fixed point of the activation function. A fixed point is further related with the equilibrium state modeling in machine learning (usually the zero gradient of the loss function in practice). 
\subsection{Comparison with Others}
We compared the proposed STL with other activation functions. And the result is summarized in Table~\ref{tab:table1}. In the terms of the desired six properties, STL can satisfy all of them, showing its advantages as activation function.

Be aware that there might be other desired properties as an activation function. But so far as we known, these six properties are proper and can be used as metrics for evaluating activation functions.

\section{Experiments}
In this section, we numerically show the advantage of the proposed STL activation function in several well-known neural networks on the CIFAR dataset. We compare STL with ReLU~\cite{Nair2010}, Mish~\cite{Misra2020} and Serf~\cite{Nag2023}. In the future, we will compare STL with more activation functions.

\subsection{CIFAR-10}
On CIFAR-10, we compared the proposed STL with ReLU, Mish and Serf in different neural networks, including SqueezeNet, Resnet-50, WideResnet-50-2, ShuffleNet-v2, ResNeXt-50, Inception-v3, DenseNet-121, MobileNet-v2, and EfficientNet-B0. In these networks, we only change the activation function. The top-1 \% accuracy values are shown in Table~\ref{tab:table2}. The results confirm that the proposed STL establishes a new state-of-the-art activation function. 
\begin{table}[!t]
	\caption{Comparison the activation functions in CIFAR-10 \label{tab:table2}}
	\centering
	\begin{tabular}{c|c|c|c|>{\columncolor{lightgray}}c}
		\hline
		Methods & ReLU &Mish &Serf & STL\\
		\hline
		SqueezeNet &84.14 &85.98 &86.32 & 86.56\\
		Resnet-50 & 86.54 &87.03 & 88.07 & 88.32\\
		WideResnet-50-2 &86.39 &86.57 &86.73 & 86.98\\
		ShuffleNet-v2 &83.93 &84.07 &84.55 & 84.86\\
		ResNeXt-50 & 87.25& 87.97 &88.49 &88.67\\
		Inception-v3 & 90.93 &91.55 &92.89 &92.93\\
		DenseNet-121 &88.59 &89.05& 89.07& 89.76\\
		MobileNet-v2 &85.74 &86.39& 86.61& 86.93\\
		EfficientNet-B0 (Swish)& 78.26 &78.02 &78.41 &78.91\\
		\hline
	\end{tabular}
\end{table}

\subsection{CIFAR-100}
On CIFAR-100, we compared the proposed STL with ReLU, Mish and Serf in different neural networks, including Resnet-164, WideResnet-28-10, DenseNet-40-12, and Inception-v3. In these networks, we only change the activation function. The top-1 \% accuracy values are shown in Table~\ref{tab:table3}. The results confirm that the proposed STL establishes a new state-of-the-art activation function.
\begin{table}[!t]
	\caption{Comparison the activation functions in CIFAR-100\label{tab:table3}}
	\centering
	\begin{tabular}{c|c|c|c|>{\columncolor{lightgray}}c}
		\hline
		Methods &ReLU &Mish &Serf &STL\\
		\hline
		Resnet-164 &74.55 &75.02 &75.13& 75.32\\
		WideResnet-28-10 &76.32 &77.03 &77.54 &77.83\\
		DenseNet-40-12 &73.68 &73.91 &74.16 &74.67\\
		Inception-v3 & 71.54 &72.38 &72.95&72.98\\
		\hline
	\end{tabular}
\end{table}

\subsection{Running Time} 
We sampled 20,000 random numbers in the interval $[-10000,10000]$ and evaluate their values with $ReLU$, $softsign$ and STL, respectively. We compare STL with these two functions because they have low computation cost. The running time for these function is 0.0054, 0.0054 and 0.0060 seconds, respectively. The proposed STL is implemented wit the naive $\log_2$ function without the numerical acceleration in Eq.~\ref{eq:acc} which can further boost the computational performance.

The $\log_2$ is available in most of programming languages such as python, C++ and MATLAB. Therefore, the proposed STL function can be easily implemented in these languages.
\section{Conclusion}
In this paper, we propose a new activation function for neural networks named the signed and truncated logarithm (STL) function. We have found that this function has several advantages over other activation functions.

Firstly, the STL function is odd, which means that it does not introduce any bias in itself. This is important because it ensures that the neural network is not skewed towards one direction or the other. 

Secondly, the function is monotonic, meaning that the relative order of the input is preserved in the output. This property is important in image recognition and classification, as well as in other applications such as speech recognition and natural language processing. 

Thirdly, the function is differentiable, which shows that it is numerically stable. This means that it can be used in a wide range of applications without encountering numerical instability issues. The differentiability property allows for efficient training of neural networks.

Fourthly, the function has an unbounded value range, which means that it can distinguish between two very large inputs. This property is particularly useful in applications where the input values can vary widely, such as in financial modeling or scientific simulations.

Fifthly, the function has a continuous gradient, which shows that it is stable with respect to the gradient. This property is important in deep neural networks, where the gradients can become unstable during training. The continuous gradient of the STL function helps ensure that the gradients remain stable, leading to faster and more stable convergence during training.

Finally, the STL function and its gradient functions can be efficiently computed in programming languages. This makes it easy to implement the function in existing neural network frameworks and to use it in a wide range of applications.

In conclusion, the STL function has significant advantages over other activation functions in neural networks. Its properties of oddness, monotonicity, differentiability, unbounded value range, continuous gradient, and computational efficiency make it an excellent choice for a wide range of applications. We believe that the STL function will play an important role in neural networks and machine learning~\cite{chenouard:2014,gong2009symmetry,Lewis2019,zhao2023survey,Gong2012,Brown2020,gong2013a,Yu2019,Gong:2014a,Yin2019a,gong:phd,Yu2022a,gong:gdp,Guo2022,gong:cf,Zong2021,gong:Bernstein,Ezawa2023,Gong2017a,Tang2021a,Gong2018,Gong2018a,Yu2020,GONG2019329,Sancheti2022,Gong2019a,Tang2021,Gong2019,Yin2019b,Gong2022,Yin2020,Gong2020a,Jin2022,Gong2021,Tang2022,Gong2021a,Tang2022a,Tang2023,Gong2022,Tang2023a,Xu2023,Han2022,Gong2023a,Scheurer2023,Gong2023,Zhang2023b}.
\bibliographystyle{IEEEtran}
\bibliography{IEEEabrv,../../IP}

\vfill

\end{document}